\documentclass[10pt,journal,compsoc]{IEEEtran}

\usepackage[nocompress]{cite}
\usepackage{graphicx}
\usepackage{amsmath}
\usepackage{amssymb}
\usepackage{booktabs}
\usepackage[pagebackref,breaklinks,colorlinks]{hyperref}
\usepackage[capitalize]{cleveref}
\crefname{section}{Sec.}{Secs.}
\Crefname{section}{Section}{Sections}
\Crefname{table}{Table}{Tables}
\crefname{table}{Tab.}{Tabs.}
\usepackage[utf8]{inputenc} 
\usepackage[T1]{fontenc}    
\usepackage{hyperref}       
\usepackage{url}            
\usepackage{booktabs}       
\usepackage{amsfonts}       
\usepackage{nicefrac}       
\usepackage{microtype}      
\usepackage{xcolor}         

\usepackage{subfigure}
\usepackage{bigstrut}
\usepackage{multirow}
\usepackage{wrapfig}
\usepackage{epsfig}
\usepackage{tablefootnote}

\newcommand{\ie}{\textit{i.e. }}

\newcommand{\etc}{\textit{etc. }}

\definecolor{TauColor}{rgb}{0.423,0.235,0.192}
\definecolor{RedColor}{rgb}{0.8,0,0}

\usepackage{silence}
\graphicspath{{./imgs/}}

\begin{document}

\title{Weakly Supervised Semantic Segmentation via Alternate Self-Dual Teaching}

\author{Dingwen~Zhang,~\IEEEmembership{Member,~IEEE,} Wenyuan~Zeng, Guangyu~Guo,
        Chaowei~Fang,
        Lechao Cheng, \\
        Ming-Ming Cheng,~\IEEEmembership{Senior Member, ~IEEE,}
        and~Junwei Han,~\IEEEmembership{Fellow,~IEEE}
\IEEEcompsocitemizethanks{\IEEEcompsocthanksitem Dingwen Zhang, Guangyu Guo, Wenyuan Zeng, and Junwei Han are with Brain and Artificial Intelligence Laboratory, School of Automation, Northwestern Polytechnical University, Xi'an, 710072, China. (e-mail: zhangdingwen2006yyy@gmail.com, junweihan2010@gmail.com). Junwei Han is the corresponding author. \\
\IEEEcompsocthanksitem Chaowei Fang is with School of Artificial Intelligence, Xidian University, Xi’an, 710071, China. (e-mail:chaoweifang@outlook.com) \\
\IEEEcompsocthanksitem Lechao Cheng is with Zhejiang Lab, Hangzhou, 310012, China. (e-mail:chenglc@zhejianglab.com) \\
\IEEEcompsocthanksitem Ming-Ming Cheng is with College of Computer Science, Nankai University, Tianjin, TKLNDST, China. 

}
\thanks{Manuscript received April 19, 2005; revised August 26, 2015.}}

\markboth{Journal of \LaTeX\ Class Files,~Vol.~14, No.~8, August~2015}%
{Shell \MakeLowercase{\textit{et al.}}: Bare Demo of IEEEtran.cls for Computer Society Journals}

\IEEEtitleabstractindextext{%
\begin{abstract}
Weakly supervised semantic segmentation (WSSS) is a challenging yet important 
research field in vision community, 
which has attached great attention in recent years. 
In WSSS, a key problem is how to generate pseudo segmentation masks (PSMs) 
that can supervise the learning of the segmentation model. 
Existing approaches generate PSMs mainly using the information cue of 
discriminative object part, 
which would inevitably encounter the problem of missing object parts 
or involving surrounding image background 
as the whole learning process is not aware of the full object structure. 
In fact, both the discriminative object part and the full object region 
are critical for guiding the generation of the fine PSMs, 
where the former provides the local object location 
while the latter provides the full object structure. 
To bring these two information cues for WSSS, 
we build a novel end-to-end learning framework, 
called alternate self-dual teaching (ASDT), 
which is designed with a dual-teacher single-student network architecture.
The information interaction process among different network branches 
is formulated from the perspective of knowledge distillation (KD). 
Unlike the conventional KD, the knowledge of the two teacher models 
would inevitably be noisy or faulty under weak supervision.
Inspired by the Pulse Width (PW) modulation used in signal processing systems,
we introduce a PW wave-like selection signal to guide the KD process, 
which can prevent the student model from falling into trivial solutions 
caused by the imperfect knowledge from either teacher model. 
Comprehensive experiments on the PASCAL VOC 2012 and COCO-Stuff 10K 
demonstrate the effectiveness of the proposed ASDT framework, 
where new state-of-the-art results are also achieved. 
\end{abstract}

}

\maketitle

\IEEEdisplaynontitleabstractindextext

\IEEEpeerreviewmaketitle

\IEEEraisesectionheading{\section{Introduction}\label{sec:introduction}}
\IEEEPARstart{S}{emantic} segmentation is a widely studied problem in computer vision. Traditional semantic segmentation approaches require 
pixel-wise manual annotation to facilitate the learning process
~\cite{long2015fully}. 
However, pixel-wise manual annotations are usually difficult to acquire due to immense labor and time costs. In order to realize the learning process of semantic segmentation model with less annotation cost, a recent trend in this research field is to develop weakly supervised semantic segmentation (WSSS) frameworks, where the annotation might be short scribbles~\cite{tang2018regularized,huang2021scribble}, bounding boxes~\cite{papandreou2015weakly},  points~\cite{bearman2016s}, or image tags~\cite{huang2018weakly,lee2021anti}. This paper focuses on the last learning scenario.   
	
	
	\begin{figure}[t]
		\centering
		\includegraphics[width=1\linewidth]{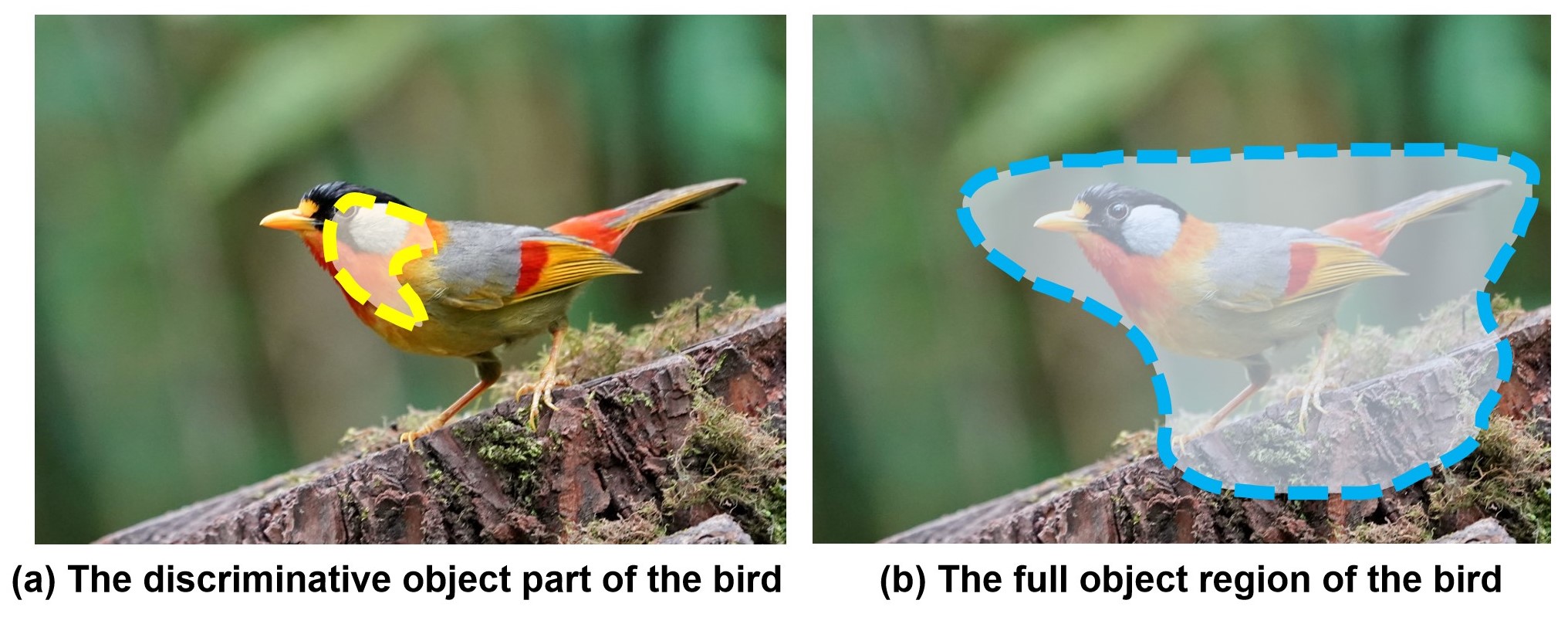}
		\caption{Revisiting the key factors for generating the high-quality PSMs for each training image. From this figure, we can observe that the discriminative object part (a) provides knowledge for localizing the object of interest, while the full object region (b) provides knowledge for revealing the whole object structure, including the beak, head, neck, wings, trunk, claws and tail of the bird. By only using the information from the discriminative object part, it is hard to infer the structure of the object entity, while using the full object region can provide complementary information.} \label{illustration}
	\end{figure}

	\begin{figure*}[t]
		\centering
		\vspace{-0.3cm}
		\includegraphics[width=1\linewidth]{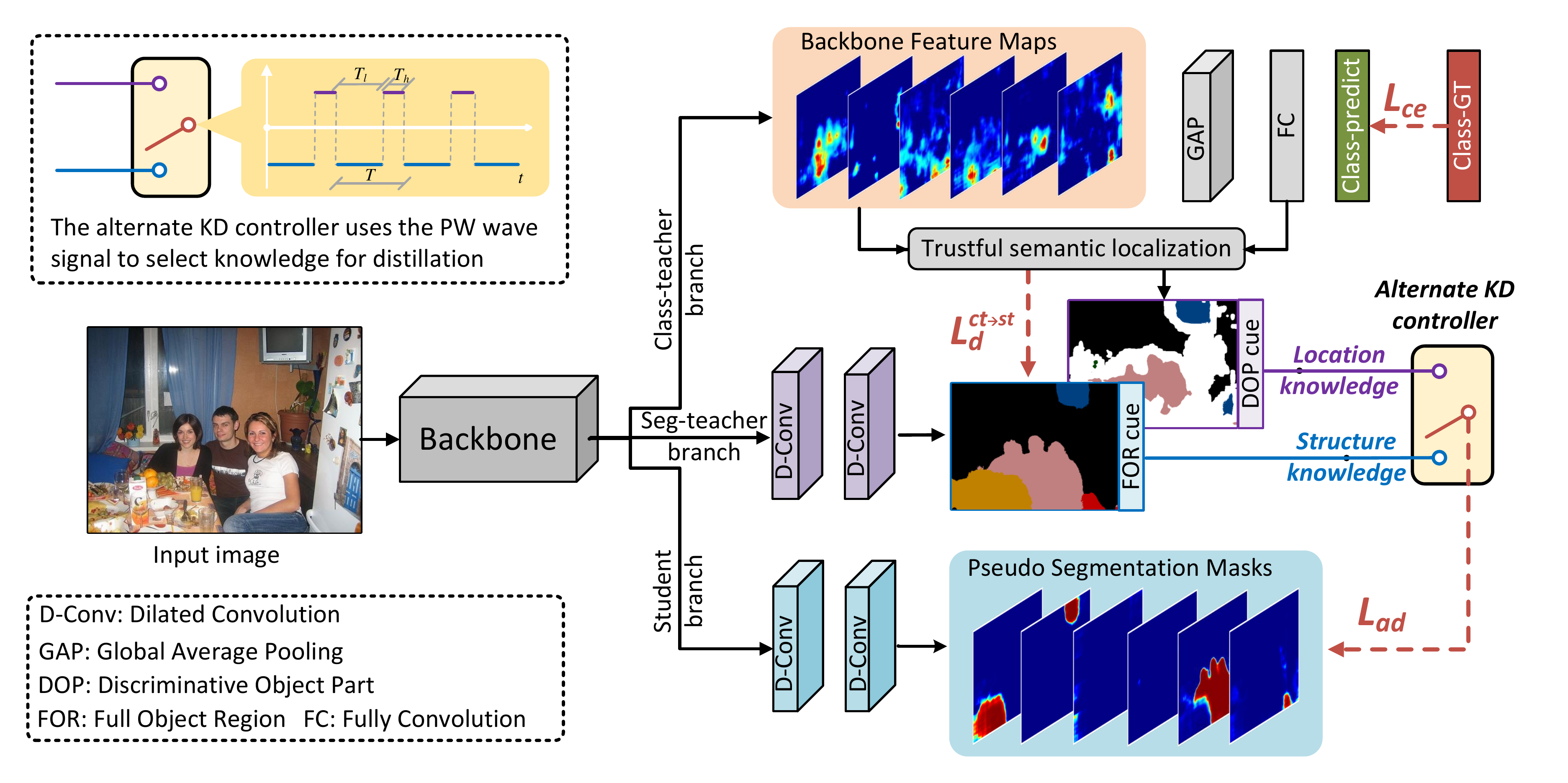}
		\caption{The overall framework of the proposed ASDT-based WSSS approach.} \label{framework}
		\vspace{-0.3cm}
	\end{figure*}
	
For addressing this task, the key problem is how to generate the pixel-wise pseudo segmentation masks (PSMs) for training images to replace the manually labeled pixel-wise annotation required for training the off-the-shelf fully supervised semantic segmentation models. 
A popular solution for PSMs generation 
\cite{fan2020employing,chang2020weakly,zhang2020causal,sun2020mining,wang2020self} 
is to use the class activation maps (CAMs)~\cite{zhou2016learning}, 
which can be extracted from the classification models.
However, as the features learned in such models are mainly used for 
the classification task rather than the segmentation task, 
generating PSMs upon these feature maps would 
inevitably have ununiform responses on object regions and 
inadequate object boundary details \cite{guo2021strengthen}. 
To address this problem, current methods 
\cite{yao2021non,zhang2020splitting,kim2021discriminative,zhang2020reliability} 
are designed to introduce an additional network model or branch that 
is dedicated to generating PSMs---we call this network model as the PSM net. Among existing works, one commonly used information cue for guiding the learning process of the PSM net is the \textit{discriminative object part} cue (see Fig. \ref{illustration} (a)). By applying different region growing-based learning strategies, these works are designed with the idea to propagate the object features from the local object part to whole object region to generate the desired PSMs. Although the discriminative object part cue can provide the key information for locating the object of interest, the model trained under such a learning paradigm is not aware of the full structure of the object. Under this circumstance, the region growing process is hard to terminate just right at the time to cover the full object without mixing up the surrounding background.

To solve this problem, this paper reveals an important yet under-studied information cue for learning the PSM net——the \textit{full object region} cue (see Fig. \ref{illustration} (b)). Specifically, the full object region indicates the image regions containing all components of an object instance but won't necessarily have accurate object boundaries. Complementary to the location knowledge provided by the discriminative object part cue, the full object region cue overviews the whole structure of the object entity, which can provide the spatial upper-bound constraints on the region growing process when implemented upon the discriminative object part. 

To this end, this paper makes an early effort to leverage both the discriminative object part cue and the full object region cue to learn to generate PSMs for WSSS. Interestingly, this process can be naturally modeled as a knowledge distillation (KD) process~\cite{ba2014deep, hinton2015distilling,gou2021knowledge} when we consider the network branches to generate the aforementioned information cues as the teacher network branches, the network branch to generate PSMs as the student network branch. Under this circumstance, the two-fold information cues generated by the teacher network branches form the knowledge that needs to be distilled to the student network branch. Benefiting from the label smoothing effect~\cite{zhang2020self}, KD has been demonstrated to be an effective way to improve the performance of the student model in different scenarios. 

Compared to the common KD processes, where the student model gets knowledge only from one teacher model, the KD process mentioned above contains two teachers in contrast. More importantly, the common KD processes are usually performed under full supervision, where the knowledge from the teacher model is perfect and accurate. Unfortunately, in WSSS, the teacher network branches can only be trained under weak supervision, making the generated knowledge not the ideal objective of the student model. Under this circumstance, constantly distilling knowledge from the teacher model to the student model like the common knowledge distillation strategies would make the student model biased to undesired trivial solutions. 


To address this issue, this paper proposes a novel alternate distillation scheme. Compared to the straightforward distillation scheme, the proposed alternate distillation scheme won't constantly distill knowledge from a certain teacher network branch. Instead, it alternatively distills knowledge from the two teacher network branches to the student branch by following a PW wave-like selection signal\footnote{Pulse-width modulation is well known in signal system. It uses a rectangular pulse wave whose pulse width is modulated resulting in the variation of the surpassing signal \url{https://en.wikipedia.org/wiki/Pulse-width_modulation}.}. In this way, the knowledge distilled from one teacher network branch can help the student branch to get rid of trivial solutions that would be derived by the other teacher branch, and vice versa. Such an alternate distillation scheme helps the student model converge to a more promising solution. Based on the alternate distillation scheme, we establish a simple but effective KD-based learning framework for WSSS, which is called alternate self-dual teaching (ASDT). As shown in Fig. \ref{framework}, ASDT is in form of a dual-teacher single-student architecture, where the two teacher network branches are designed to generate the knowledge about the \textit{discriminative object part} and \textit{full object region}, respectively. Then, the object location knowledge and structure knowledge would be selectively distilled via an alternate KD controller to guide the learning process of the student network branch, which enables the model to obtain more reliable PSMs.

To sum up, there are three-fold contributions in this work:
	\begin{itemize}
		\item By revisiting the key factors for producing high-quality PSMs, this paper reveals that two-fold information cues, i.e., the \textit{discriminative object part} and \textit{full object region}, are both critical for WSSS. Based on this, a novel end-to-end ASDT framework is built to formulate the problem from the perspective of KD.  
		
		\item Unlike common KD schemes, we propose alternate distillation scheme to work under the weak supervision. By alternately distilling knowledge under the guidance of a PW wave-like selection signal, the student network branch won't get stuck into the trivial solutions caused by the imperfect knowledge of either teacher network branch.

		\item Comprehensive experiments on two widely-used benchmarks have been implemented to evaluate the performance of the proposed approach. Experimental results demonstrate that such an easy-to-implement scheme achieves superior segmentation performance compared to the existing state-of-the-art methods. 
	\end{itemize}
\section{Related Works}
	
\subsection{Weakly Supervised Semantic Segmentation}
Weakly-supervised semantic segmentation (WSSS) under image-level supervision is the most challenging task among all the WSSS categories as it requires minimal effort for human annotation. Approaches designed in early ages operate only with the image-level supervision without using additional information ~\cite{ahn2018learning,shimoda2019self}. Most of these kinds of methods first generate pseudo-labels based on CAMs~\cite{zhou2016learning}, and then train a supervised semantic segmentation network such as DeepLab-v1~\cite{ChenPKMY14Semantic}, DeepLab-v2~\cite{chen2017deeplab}. 
While many recent approaches introduce extra information into the WSSS training process to obtain better performance~\cite{lee2019frame,liu2020leveraging}. In these methods, saliency map~\cite{borji2015salient} usually works as the most popular prior information as it can provide accurate boundary information~\cite{qi2016augmented,kolesnikov2016seed,huang2018weakly}. While some other methods introduce extra training images such as noisy images obtained from web~\cite{jin2017webly,lee2019frame}, YouTube videos~\cite{hong2017weakly}, or ImageNet (\ie 24K ImageNet)~\cite{hou2017bottom}.
In this paper, we design a novel WSSS framework only using the image-level supervision. 

\subsection{Knowledge Distillation}
Knowledge distillation~\cite{ hinton2015distilling} aims to transfer knowledge from a well-trained teacher network to a compact student network. In the classical knowledge distillation approaches, the student networks are supervised by information extracted from the teacher networks, such as predicted probabilities~\cite{ba2014deep, hinton2015distilling}, intermediate features~\cite{passalis2018learning}, \etc Differently, the self-distillation mechanism transfers knowledge within a model itself~\cite{zhang2019your,yang2019snapshot,phuong2019distillation}. For example, \cite{zhang2019your} transfers knowledge from the deeper layers of a neural network into its shallow layers. 
\cite{phuong2019distillation} proposes to guide the learning of a current network layer by the output of network layer behind it. \cite{yang2019snapshot} utilizes the information of earlier training epochs to supervise the later training epochs.  
The self-distillation mechanism has been applied in many fields like classification~\cite{zhang2019your}, weakly-supervised object detection~\cite{huang2020comprehensive}, \etc In this paper, we introduce the self-distillation mechanism into the training process of the WSSS model and propose a novel self-dual teaching strategy to facilitate an effective knowledge distillation under the weak supervision.
\section{Approach}
	
Given a training image collection $\mathcal{X}$ and the corresponding image-level label collection $\mathcal{Y}$, we build an end-to-end learning framework to predict pixel-level PSMs $\bar{\textbf{P}}$ for each training image. As shown in Fig. \ref{framework}, the learning framework first has a feature extraction backbone followed by a self-dual teaching network architecture, which contains three parallel network branches including the class-teacher branch $f_{ct}(\cdot)$, the seg-teacher branch $f_{st}(\cdot)$, and the student branch $f_{s}(\cdot)$. During training, the two teacher network branches distill two-fold gainful knowledge, including the object location knowledge (from the discriminative object part) and the object structure (from the full object region), to the student network branch. In order to enable an effective dual-teacher KD process under the weak supervision, we further develop an alternate distillation mechanism, which uses the PW wave-like selection signal to control the alteration of the knowledge in the KD process.

\begin{figure*}[t]
	\centering
	\includegraphics[width=1.0\linewidth]{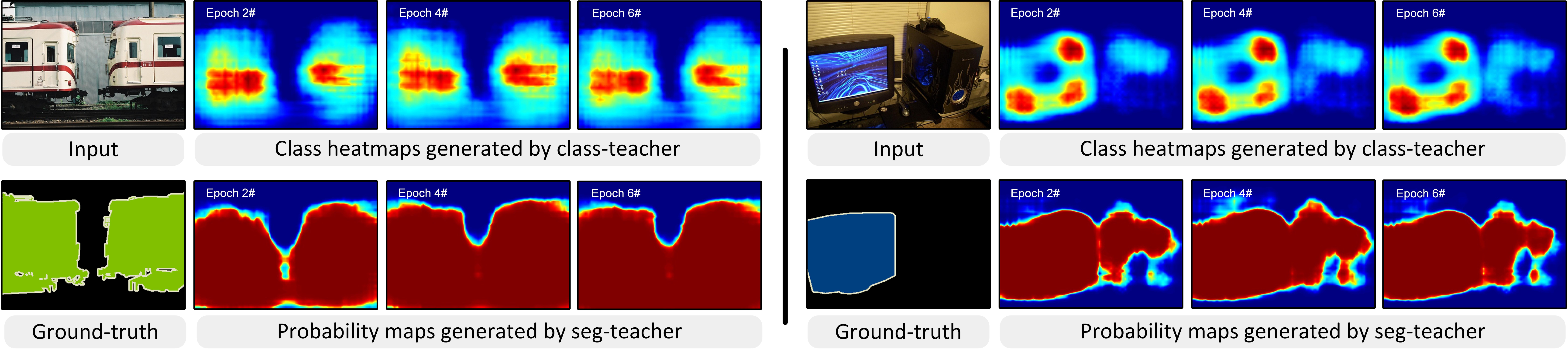}
	\caption{Examples to show different characteristics of the knowledge extracted by the class-teacher branch and seg-teacher branch. As can be seen, the class-teacher would focus on the local parts of the whole objects of interest, while the seg-teacher would segment out more visually similar regions but introduce miss-classification regions.}
	\label{inter1}
\end{figure*}

Once the model is trained, we generate the final PSM by collaborating the seg-teacher network branch and the student network branch. 
Specifically, we first combine the predictions of the seg-teacher network branch $\textbf{P}^{st}$ and the student network branch $\textbf{P}^{s}$ by:
\begin{equation}
	\bar{\textbf{P}}=\Delta\max[\textbf{P}^{st},\textbf{P}^{s}],
	\label{combine}
\end{equation}
where $\Delta\max$ indicates the element-wise maximization operation for the input matrix. As one of the common operation in this research field, we then adopt the CRF-based post-processing to obtain the final PSM prediction $\bar{\textbf{P}}$.

\subsection{Self-Dual Teaching Architecture}
\label{Self-Dual Teaching Architecture}
In the self-dual teaching architecture, the class-teacher network branch is used to generate the discriminative object part cue. In this network branch, we first perform the global average pooling (GAP) \cite{zhou2016learning} on the feature maps extracted by the feature extraction backbone $\textbf{F}$. Then, a fully-connected layer is adopted to map the pooled feature vector to the image-level classification prediction. The whole process of this network branch can be denoted as $\hat{\textbf{y}}=f_{ct}(\textbf{F},\textbf{W}_{ct})$, where $\textbf{W}_{ct}$ indicates the learnable parameters of this network branch. Under the supervision of the image-level ground-truth  $\textbf{y}=[y_1,y_2,\cdot,y_C]$, where $C$ indicates the total number of the explored classes, we learn the class-teacher by minimizing:
\begin{equation}
	\mathcal{L}_{ce}=-\sum_{c=1}^C [{y}_c \ln \hat{y}_c + (1-{y}_c) \ln (1-\hat{y}_c)].
	\label{ct}
\end{equation}
For generating the discriminative object part cue, we follow \cite{zhou2016learning} to associate the feature maps $\textbf{F}$ with the learned weights in the fully-connected layer, thus generating the class-wise heat-maps $\mathcal{H}=\{\textbf{H}_c\}_{c=1}^C$. To identify the trustful semantic location from $\mathcal{H}$, we follow \cite{zhang2020reliability,zhang2021adaptive} to generate the binary label matrix $\textbf{B}^{ct}\in \mathbb R^{h\times w\times (C+1)}$ and the reliability mask $\textbf{R}^{ct}\in \mathbb R^{h\times w}$ based upon $\mathcal{H}$, where the reliability mask $\textbf{R}^{ct}$ can help screen the sparse yet highly activated noisy locations produced by the CAM for a more stable procedure in the subsequent distillation process.

The seg-teacher is a segmentation-oriented network branch with parameters $\textbf{W}_{st}$ and output $\textbf{P}^{st}$. This network branch is formed by two $3 \times 3$ dilated convolutional layers with the dilation rate of 12 and a softmax operation. This branch is trained directly under the guidance of the class-teacher: 
\begin{equation}
		\mathcal{L}_{d}^{ct \rightarrow st}=\sum_i\sum_{c\in \mathcal{C}^+}r_i^{ct} b_{c,i}^{ct}\log(p^{st}_{c,i})+\mathcal{L}_{str},
	\label{d1}
\end{equation}
where $r_{i}^{ct}$, $b_{c,i}^{ct}$, and $p^{st}_{c,i}$ are the elements in $\textbf{R}^{ct}$, $\textbf{B}^{ct}$, and $\textbf{P}^{st}$, respectively. $\mathcal{C}^+$ is the collection of presented object categories, which is obtained according to the image-level annotation. As used in \cite{huang2018weakly,zhang2020reliability}, $\mathcal{L}_{str}$ is an energy-based term to help explore the spatial and low-level appearance dependency in the knowledge distillation process.

As such a learning process makes the model search similar visual patterns from the surrounding regions of the local object part, it would inevitably introduce the background context in the generated segmentation masks (see the bottom row of Fig. \ref{inter1}), which happen to be the full object region explored by this work. Once this  network branch is trained, we employ the CRF-based post-processing upon $\textbf{P}^{st}$ to obtain $\textbf{B}^{st}$, which encodes the object structure knowledge for learning the student branch.

The student network branch $f_{s}(\cdot)$ has the same architecture as $f_{st}(\cdot)$, but distinct network parameters $\textbf{W}_s$. It is trained based on the knowledge distilled from the above two teacher branches under the alternate distillation mechanism (see the details in the next subsection) and predicts the probability maps $\textbf{P}^s=f_s(\textbf{F}|\textbf{W}_s)$ with $C+1$ channels. 

\subsection{Alternate Distillation Mechanism}
\label{Alternative Distillation Mechanism}
In the proposed self-dual teaching architecture, the two teacher network branches have different network designs and are learned in different manners---The class-teacher $f_{ct}(\cdot)$ is learned under the image-level weak supervision, while the seg-teacher $f_{st}(\cdot)$ is learned under the self-produced supervision. Under this circumstance, the guidance knowledge generated by the two teacher models would have different properties as well: The class-teacher would overweight the classification results of the image so that it focuses more on the local discriminative object parts but ignores the less discriminative yet indispensable object parts (see top row of Fig. \ref{inter1}). On the contrary, the seg-teacher would overweight the classification results on each pixel to pursuit for high segmentation performance, so that it can segment out the large portion of the object regions but would miss-classify the image regions with the large intra-class variation or inter-class similarity (see bottom row of Fig. \ref{inter1}).

From the above discussion, we can see that both of the teacher models have their own merits to distill helpful knowledge but neither of them is perfectly correspond to the desired PSM. To this end, directly distilling knowledge from either the class-teacher or the seg-teacher would lead the student network branch to fall into undesired trivial solution. To solve this problem, we propose a novel alternate distillation mechanism. Under this mechanism, the knowledge used to guide the learning process of the student network branch is alternatively selected from the class-teacher and the seg-teacher, which is controlled by a PW wave-based selection signal (see Fig. \ref{framework}). Specifically, the alternate distillation loss is defined as:
\begin{equation}
    \mathcal{L}_{ad}=\Lambda_t(\mathcal{L}_{d}^{ct \rightarrow s}(\textbf{P}^s,[\textbf{B}^{ct},\textbf{R}^{ct}]),\mathcal{L}_{d}^{st\rightarrow s}(\textbf{P}^s,\textbf{B}^{st})|T,\tau),
		\label{ad}
\end{equation}
where $\Lambda_t(\cdot)$ indicates the alternate KD controller, which alternately selects the guiding knowledge according to the PW wave-like signal along iterations. $T$ and $\tau$ are the hyper-parameters for identifying the PW wave signal, where $T=T_h+T_l$ indicates the alternation period width while $\tau=T/T_h$. Here $T_h$ and $T_l$ are the length of the high level signal (the signal to distill object location knowledge) and the low level signal (the signal to distill object structure knowledge) in one alternation period width. The PW wave signal can be referred to in the left-top corner of Fig. \ref{framework}. The $\mathcal{L}_{d}^{ct \rightarrow s}$ and $\mathcal{L}_{d}^{st \rightarrow s}$ are defined as:
\begin{equation}
		\mathcal{L}_{d}^{ct \rightarrow s}=\sum_i\sum_{c\in \mathcal{C}^+}r_i^{ct} b_{c,i}^{ct}\log(p^s_{c,i})+\mathcal{L}_{str}, 
	\label{d2}
\end{equation}
\begin{equation}
		\mathcal{L}_{d}^{st \rightarrow s}=\sum_i\sum_{c\in \mathcal{C}^+} b^{st}_{c,i}\log(p^s_{c,i})+\mathcal{L}_{str}. 
	\label{d3}
\end{equation}	
As can be seen, different from the conventional knowledge distillation methods that use the soft label as the guidance knowledge, we use the hard label instead. This is because the conventional KD methods work under the supervised learning scenario, whereas the KD process performed by our framework is under weak supervision. According to ~\cite{ji2020knowledge}, when the teacher models cannot provide perfect knowledge to the student model as in our case, using the soft label would introduce more disturbing knowledge that hurts the learning of the student model. At this moment, using the hard labels can sometimes correct teachers' wrong prediction and thus obtain better performance. 





\section{Experiments}
\label{sec:exp}

\textbf{Dataset and Evaluation Metric.} Experiments are conducted on the Pascal VOC 2012~\cite{everingham2010pascal} dataset and COCO-Stuff 10K~\cite{lin2014microsoft}. On the Pascal VOC 2012, following the common practice~\cite{zhang2020reliability,lee2021anti}, we train our model on 10582 images where the extra images and annotations are from~\cite{hariharan2011semantic}. We report the evaluation results on 1449 validation images and 1456 test images. We adopt mean intersection-over-union (mIoU) as the evaluation metric.

\textbf{Implementation Details.}
When implementing our approach, we use \cite{zhang2020reliability} as our baseline model. 
For a training image, we resize it with a random ratio from (0.7, 1.3), and then apply a random flip. Finally, it is normalized and  cropped into $321\times 321$ images. For the PW wave signal, we set $T=150$ and $\tau=5$. To generate reliable pseudo labels, the scale ratio of multi-scale CAM is set to \{0.5, 1, 1.5, 2\}. During testing, Dense CRF is used as post-processing. We use the SGD optimizer~\cite{bottou2010large} with a momentum of 0.9 and a weight decay of $10^{-5}$. The learning rate is 0.0007. We train the network for 8 epochs with a batch size of 4. We set $\sigma_{D}=15$ and $\sigma_{I}=100$ as \cite{zhang2020reliability}.
We set the momentum and weight decay of the SGD optimizer as 0.9 and $5\times 10^{-4}$, respectively. The initial learning rate is 0.0025 and is decreased according to the polynomial decay policy with a power of 0.9. We train the segmentation network for 20000 iterations with a batch size of 10. After predicting the PSM for each training image, we apply two different segmentation networks to produce the final segmentation results. In the first segmentation network, we follow~\cite{zhang2020reliability,chen2020weakly,zhang2020causal,lee2021anti} to utilize a PyTorch implementation of the DeepLab-v2~\cite{chen2017deeplab} with ResNet-101~\cite{he2016deep} or ResNet-38~\cite{wu2019wider} backbone \footnote{\url{https://github.com/kazuto1011/deeplab-pytorch}}. In the second segmentation network, we follow~\cite{li2021pseudo} to use PSPnet~\cite{zhao2017pyramid} with Res2Net101~\cite{gao2019res2net} backbone.
The experiments are implemented by Pytorch~\cite{Paszke2017at} on an NVIDIA GTX 1080Ti GPU. 

\begin{table}[t]
\footnotesize
  \centering
  \caption{Comparison with the state-of-the-art approaches on PASCAL VOC 2012 \textit{val} and \textit{test} sets. the supervision information (Sup.) includes: F (full pixel-level supervision), I (image-level supervision), B (bounding box-level supervision), S (scribble-level supervision), SA (saliency maps).}
  \renewcommand{\arraystretch}{1.0}{
  \setlength{\tabcolsep}{1.8mm}{
    \begin{tabular}{l|c|cccc}
    \hline
    Method & Pub.  & Backbone & Sup.  &  val  &  test \bigstrut\\
    \hline
    DeepLab-v1~\cite{ChenPKMY14Semantic} & ICLR15 & VGG-16 & F     & 67.6  & 70.3  \bigstrut[t]\\
    Deeplab-v2~\cite{chen2017deeplab} & PAMI17 &  ResNet-101 & F     & 76.8  & 79.7  \bigstrut[b]\\
    \hline
    WSSL~\cite{papandreou2015weakly} & ICCV15 & VGG-16 & B     & 60.6  & 62.2  \bigstrut[t]\\
    BBAM~\cite{lee2021bbam} & CVPR21 & ResNet-101 & B     & 73.7  & 73.7  \\
    Oh \textit{et al}.~\cite{oh2021background} & CVPR21 & ResNet-101 & B     & 74.6  & 76.1  \bigstrut[b]\\
    \hline
    KernelCut~\cite{tang2018regularized} & ECCV18 &  ResNet-101 & S     & 75.0  & - \bigstrut[t]\\
    BPG~\cite{wang2019boundary} & IJCAI19 &  ResNet-101 & S     & 76.0  & - \bigstrut[b]\\
    \hline
    Fan \textit{et al}.~\cite{fan2020employing} & ECCV20 &  ResNet-101 & I, SA & 67.2  & 66.7  \bigstrut[t]\\
    Lee \textit{et al}.~\cite{lee2019frame} & ICCV19 &  ResNet-101 & I, SA & 66.5  & 67.4  \\
    MCIS \textit{et al}.~\cite{sun2020mining} & ECCV20 & ResNet-101 & I, SA & 67.7  & 67.5  \\
    ICD~\cite{fan2020learning} & CVPR20 & ResNet-101 & I, SA & 67.8  & 68.0  \\
    LIID~\cite{liu2020leveraging} & PAMI20 & ResNet-101 & I, SA & 67.8  & 68.3  \\
    Li  \textit{et al}.~\cite{li2020group} & AAAI21 & ResNet-101 & I, SA & 68.2  & 68.5  \\
    Yao \textit{et al}.~\cite{yao2021non} & CVPR21 & ResNet-101 & I, SA & 68.3  & 68.5  \\
    AuxSegNet~\cite{xu2021leveraging} & ICCV21 & ResNet38 & I, SA & 69.0  & 68.6  \\
    SPML~\cite{ke2021universal} & ICLR21 & ResNet-101 & I, SA & 69.5  & 71.6  \\
    EDAM~\cite{wu2021embedded} & CVPR21 & ResNet-101 & I, SA & 70.9  & 70.6  \\
    DRS~\cite{kim2021discriminative} & AAAI21 & ResNet-101 & I, SA & 71.2  & 71.4  \bigstrut[b]\\
    \hline
    ICD~\cite{fan2020learning} & CVPR20 & ResNet-101 & I     & 64.1  & 64.3  \bigstrut[t]\\
    IRN~\cite{ahn2019weakly} & CVPR19 & ResNet50 & I     & 63.5  & 64.8  \\
    SSDD~\cite{shimoda2019self} & ICCV19 & ResNet-38 & I     & 64.9  & 65.5  \\
    SEAM~\cite{wang2020self} & CVPR20 & ResNet-38 & I     & 64.5  & 65.7  \\
    Chang \textit{et al}.~\cite{chang2020weakly} & CVPR20 & ResNet-101 & I     & 66.1  & 65.9  \\
    RRM~\cite{zhang2020reliability} & AAAI20 & ResNet-101 & I     & 66.3  & 66.5  \\
    BES~\cite{chen2020weakly} & ECCV20 & ResNet-101 & I     & 65.7  & 66.6  \\
    Ru \textit{et al}.~\cite{ru2021learning} & IJCAI21 & ResNet101 & I     & 67.2  & 67.3  \\
    CONTA~\cite{zhang2020causal} & NIPS20 & ResNet-101 & I     & 66.1  & 66.7  \\
    ECS-Net~\cite{sun2021ecs} & ICCV21 & ResNet38 & I     & 66.6  & 67.6  \\
    CPN~\cite{zhang2021complementary} & ICCV21 & ResNet38 & I     & 67.8  & 68.5  \\
    AdvCAM~\cite{lee2021anti} & CVPR21 & ResNet-101 & I     & 68.1  & 68.0  \\
    PMM~\cite{li2021pseudo} & ICCV21 & ResNet-38 & I     & {68.5 } & {69.0 } \\
    Ours  & -     & ResNet-101 & I     & {68.5 } & 68.4\tablefootnote{\url{http://host.robots.ox.ac.uk:8080/anonymous/HMPFSZ.html}}  \bigstrut[b]\\
    Ours  & -     & ResNet-38 & I     & \textbf{69.7 } & \textbf{70.1}\tablefootnote{\url{http://host.robots.ox.ac.uk:8080/anonymous/46YUZO.html}}  \bigstrut[b]\\
    \hline
    LIID~\cite{liu2020leveraging} & PAMI20 & Res2Net-101 & I & 69.4  & 70.4 \\
    PMM~\cite{li2021pseudo} & ICCV21 & Res2Net-101 & I     & 70.0  & 70.5  \bigstrut[t]\\
    Ours  & -     & Res2Net-101 & I     & \textbf{71.0 } & \textbf{71.0 }\tablefootnote{\url{http://host.robots.ox.ac.uk:8080/anonymous/R9XFGZ.html}} \bigstrut[b]\\
    \hline
    \end{tabular}%
    }}
  \label{tab:stoa}%
\end{table}%

\subsection{Comparison with Stat-of-the-art Methods}
\label{sec:expstoa}
\begin{figure*}[t]
	\centering
	\includegraphics[width=1\textwidth]{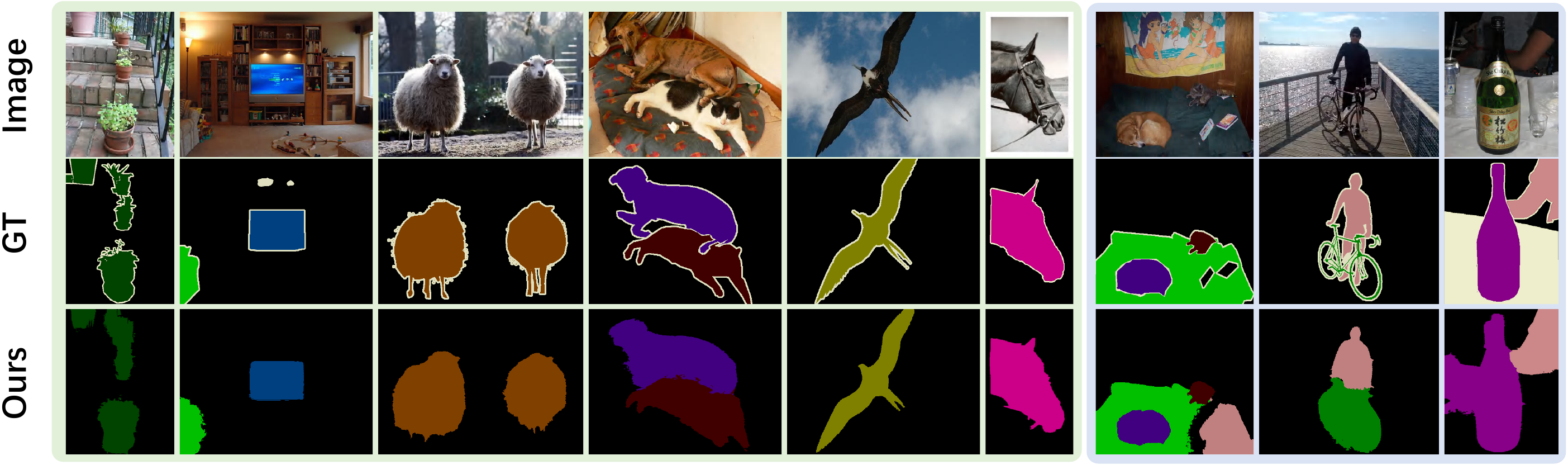}
	\caption{Some segmentation results of our method, including six successful examples (left) and three failed examples (right).} 
	\label{visual}
\end{figure*}

\textbf{Pascal VOC 2012.} In Table~\ref{tab:stoa}, we compare our proposed method with state-of-the-art methods that are learned with various levels of annotation, including fully supervised masks (F), bounding boxes (B), scribbles (S), or image class labels (I), with and without extra data. We report the performance of the final segmentation model. According to the experimental results, our method obtains mIoU of 68.5\% and 68.4\%, which is 2.2\% and 1.9\% higher than mIoU of our baseline model RRM on validation and test set, respectively. Compared to the start-of-the-art method without using saliency maps, i.e., PMM, our method achieves 1.2\% and 1.1\% higher mIoU scores on the validation and test set, respectively, when using ResNet-38 as the segmentation network backbone. When using the Res2Net-101 as the segmentation network backbone, our approach still outperforms PMM on both validation and test set. It is worth mentioning that the ResNet-38 backbone used in our experiment as well as those in previous WSSS works is not the original ResNet model \cite{he2016deep}. It is actually heavier than the ResNet-101 model (104.1M parameters V.S. 40.5M parameters). 
Some visualization examples of the proposed approach, including six successful examples and three failed examples, are shown in Fig. \ref{visual}. In particular, we can observe from three failed examples that it is still difficult for our method to distinguish group objects overlapping with each other.

\textbf{COCO-Stuff 10k.} Apart from PASCAL VOC 2012, we also provide results on COCO-Stuff 10K~\cite{lin2014microsoft} dataset. Following EDAM~\cite{wu2021embedded}, we selected 9000 images that belong to the 20 categories of PASCAL VOC for training and set pixels of other categories as background. The segmentation model is DeepLab-v2 with ResNet-101 backbone. As shown in Table~\ref{tab:cocostuff}, our ASDT brings 0.6\% performance gains over baseline RRM. Moreover, our ASDT outperforms EDAM, which uses saliency maps in post-processing. Those results demonstrate the proposed ASDT can perform well in more complex scenarios.

\begin{table}[t]
\centering
\caption{Experimental results of ResNet101 backbone on COCO-Stuff 10K. The supervision information (Sup.) includes: F (full pixel-level supervision), I (image-level supervision), SA (saliency maps). $^\dagger$ denotes the results from EDAM~\cite{wu2021embedded}} 
\setlength{\tabcolsep}{5mm}{
\begin{tabular}{l|cc}
\toprule
Method & Sup.  & test \\
\midrule
Deeplab-v2~\cite{chen2017deeplab}$^\dagger$ & F     & 55.9 \\
\midrule
EDAM~\cite{wu2021embedded}$^\dagger$ & I, SA & 51.4  \\
RRM~\cite{zhang2020reliability} & I     & 51.4  \\
Ours  & I     & \textbf{52.0 } \\
\bottomrule
\end{tabular}%
}
\label{tab:cocostuff}%
\end{table}%

\subsection{Experiments with Extra Data}
We also follow existing works~\cite{li2019attention,sun2020mining,wu2021embedded} to evaluate the weakly supervised learning performance of the proposed approach by using extra weakly labeled training data. Specifically, the experiments are carried out by using the additional single-label images from the Caltech-256 dataset~\cite{griffin2007caltech}, where around 4000 images are selected to align with the object categories presented in PASCAL VOC 2012.
The experimental comparison results are reported in Table~\ref{tab:extra}. We can observe our ASDT outperforms all other methods, even though some of them use extra saliency maps \cite{yao2021non,xu2021leveraging,liu2020leveraging}. Note that MCIS~\cite{sun2020mining} uses 20k extra images from Caltech-256 and ImageNet.

\subsection{Ablation Study}
\label{Ablation Study}

In Table~\ref{tab:ablation}, we first compare different self-distillation strategies on the validation set of the PASCAL VOC 2012 benchmark. We report the mIoU performance of the seg-teacher, student, and the final combined prediction. The concrete settings are described as follows:

\begin{itemize}
	\item In experiment \textbf{(1)}, the model only contains a class-teacher branch and a student branch. The network parameters are trained with $\mathcal{L}_{ce}$ and $\mathcal{L}_d^{ct \rightarrow st}$.
	
	\item In experiment \textbf{(2)}, the model only contains a class-teacher branch, a seg-teacher branch, and a student branch, learned by $\mathcal{L}_{ce}$, $\mathcal{L}_d^{ct \rightarrow st}$, and $\mathcal{L}_d^{st \rightarrow s}$, respectively.
	\item In experiment \textbf{(3)}, the model is similar to the experiment \textbf{(2)}. The only difference is that the supervision signal for training the student branch is obtained by performing CRF on $\max(\textbf{P}^{ct},\textbf{P}^{st})$ instead of $\textbf{B}^{st}$.
	
	\item In experiment \textbf{(4)}, the model performs CRF on $\text{mean}(\textbf{P}^{ct},\textbf{P}^{st})$ to train the student branch. Other settings are kept the same with the experiment \textbf{(3)}.   
\end{itemize}

	\begin{table}[t]
\centering
\caption{Comparison with the state-of-the-art approaches on PASCAL VOC 2012 with extra simple single-label images from Caltech-256~\cite{griffin2007caltech}. The supervision information (Sup.) includes: I (image-level supervision), SA (saliency maps).}
\setlength{\tabcolsep}{4mm}{
 \begin{tabular}{l|c|cc}
\toprule
Method & Sup.  & val   & test \\
\midrule
MCNN~\cite{tokmakov2016weakly} & I     & -     & 36.9  \\
MIL-seg~\cite{pinheiro2015image} & I     & 42.0  & 40.6  \\
AttnBN~\cite{li2019attention} & I, SA & 66.1  & 65.9  \\
MCIS~\cite{sun2020mining} & I, SA & 67.1  & 67.2  \\
EDAM~\cite{wu2021embedded} & I, SA & \textbf{72.0 } & 71.4  \\
\midrule
Ours  & I     & \textbf{72.0 } & \textbf{71.9 }\tablefootnote{\url{http://host.robots.ox.ac.uk:8080/anonymous/ICEZ2N.html}} \\
\bottomrule
\end{tabular}%
}
\label{tab:extra}%
\end{table}%

From the reported results, we can observe that using the single-teacher model can already achieve a good segmentation performance. However, it is nontrivial to further introduce dual-teaching architecture to learn the student network branch. In particular, as shown in experiment \textbf{(2)}, the student branch only achieves 30.4\% mIOU when it is supervised only by the seg-teacher, which reveals that the performance of a segmentation-oriented branch is seriously affected by the errors from the supervision signal. In fact, the seg-teacher itself can only generate segmentation results with 62.3\% mIoU, which indicates that it would produce lots of miss-classification and be far from a perfect teacher model, especially at the early phase of the learning procedure. Under this circumstance, the conventional direct distillation strategy cannot learn a good student model. In the experiment \textbf{(3)} and \textbf{(4)}, two naive ways are tried to combine the two different teacher network branches: using element-wise mean or element-wise max for generating the fused supervision maps. Although the mIoU of the student increases from 30.4\% to 40.1\% and 40.0\%, respectively, such results are still unsatisfactory.
Compared to the above four experiments, in experiment \textbf{(5)}, we adopt the newly proposed distillation mechanism which alternatively chooses the two different teacher branches. In this case, the performance of the student branch reaches 63.8\%, and the mIoU of the combined prediction improves to 64.0\%. The superiority of our learning mechanism is its capacity in correcting the error supervision of the seg-teacher branch in the early learning phase while integrating diverse supervision signals in the late learning phase.

\begin{table}[t]
	\centering
	\caption{Analysis about different distillation strategies on the PASCAL VOC 2012 validation set. The reported results correspond to the PSMs generated by the learning framework. }
			\begin{tabular}{c|l|ccc}
				\toprule
			 &{Distillation Strategy} & $\textbf{P}^{st}$ & $\textbf{P}^{s}$ & $\bar{\textbf{P}}$ \\
				\midrule
			I& { Single class-teacher} & -     & 62.6  & - \\
			II&{ Single seg-teacher } & 62.3  & 30.4  & 48.5  \\
			III& { Naive dual teacher (max)} & 6I.4  & 40.1  & 53.2  \\
			IV&	 { Naive dual teacher (mean)} & 62.3  & 40.0  & 53.6  \\
			 V&{ Alternate dual teacher} & 63.8  & 63.8  & 64.0  \\
				\bottomrule
			\end{tabular}%
	\label{tab:ablation}%
\end{table}%


\subsection{Analysis on the Pulse Width Modulation}

\begin{figure}[t]
	\centering
	\includegraphics[width=1\linewidth]{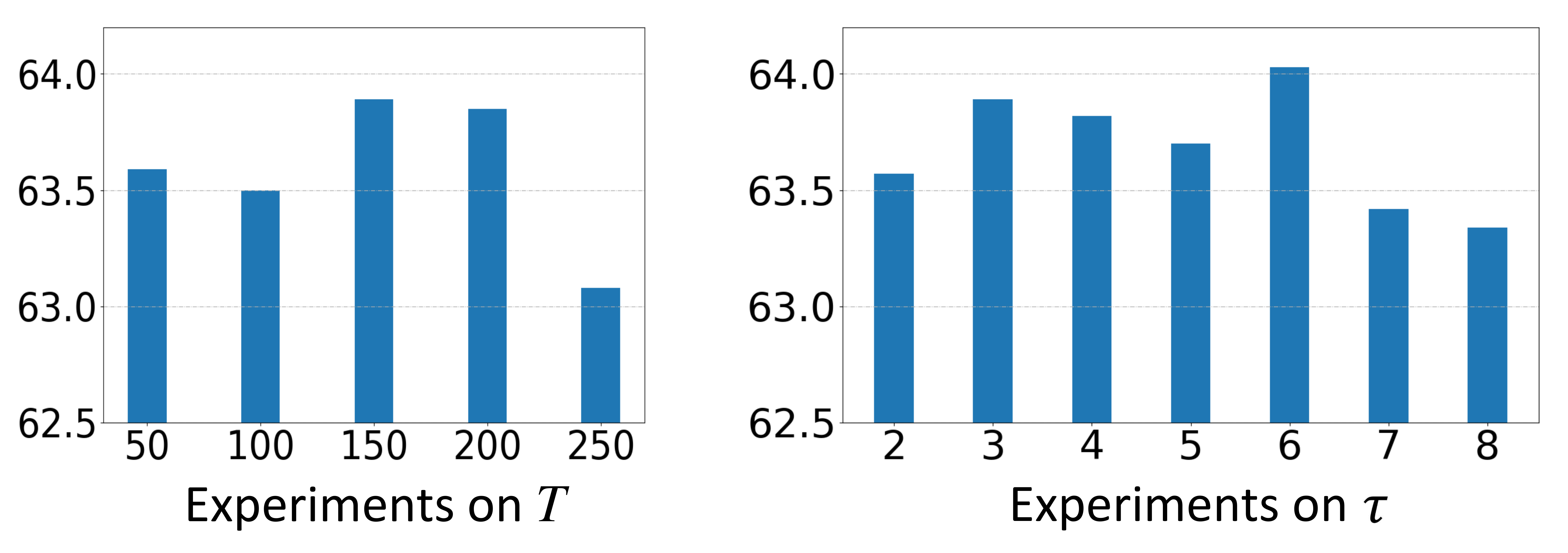}
	\caption{Performance of using different values for $T$ and $\tau$ on the PASCAL VOC 2012 validation set. The reported results correspond to the PSMs generated by the learning framework.}
	\label{para}
\end{figure}

The proposed alternate distillation mechanism is determined by a pulse width modulation, and the shape of the PW wave is controlled by alternation period width $T$ and alternation proportion $\tau$. In Fig.~\ref{para}, we analyze how these two hyper-parameters affect the performance of the proposed alternate self-dual teaching model.

\textbf{Alternation period width $T$:} This parameter determines how many iterations used in a single self-dual teaching period. As shown in the left figure of Fig.~\ref{para}, we compared five different period widths. From the experimental results, we can observe that different period widths would cause 1 mIoU effect on the quality of the generated PSMs .

\textbf{Alternation proportion $\tau$:} This parameter determines the proportion of the class-teacher and the seg-teacher within a self-dual teaching period. As shown in the right of Fig.~\ref{para}, varying $\tau$ to different values 
would have a stronger influence than the $T$ and setting $\tau=6$ obtains relatively better performance than other values.

\section{Conclusion}
\label{sec:con}
In this paper, we build a novel end-to-end WSSS framework for 
generating the PSMs under the weak image-level supervision. By introducing the full object region cue as a complement to the commonly used discriminative object part cue, a novel dual teacher network architecture is established, where KD processes are implemented between different network branches. To facilitate an effective KD under the weak supervision, we further propose the alternate distillation scheme, in which the KD process of the two-fold knowledge is controlled by a PW wave-like selection signal. 
Comprehensive experiments demonstrate the effectiveness of the proposed approach. In the future, we will improve the ASDT mechanism and apply it to a wide range of weakly supervised learning tasks, such as weakly supervised object detection \cite{zhang2021weakly} and instance segmentation \cite{liu2020leveraging}.

\ifCLASSOPTIONcaptionsoff
  \newpage
\fi



\bibliographystyle{IEEEtran}
\bibliography{IEEEabrv,WSSS_ASDT_arxiv}
%
%

\end{document}